\documentclass[10pt,twocolumn,letterpaper]{article}

\usepackage{iccv}
\usepackage{authblk}
\usepackage{times}
\usepackage{epsfig}
\usepackage{graphicx}
\usepackage{amsmath}
\usepackage{amssymb}
\usepackage{enumitem}
\usepackage{multirow}
\usepackage{booktabs}

\usepackage[T1]{fontenc}
\usepackage[utf8]{inputenc}
\usepackage[english]{babel}
\usepackage[autostyle, english=american]{csquotes}
\MakeOuterQuote{"}

\usepackage[breaklinks=true,bookmarks=false]{hyperref}

\iccvfinalcopy 

\def\iccvPaperID{****} 

\ificcvfinal\fi

\begin{document}

\title{The Solution for the ICCV 2023 Perception Test Challenge 2023 - Task 6 - Grounded videoQA}


\author{
  Hailiang Zhang$^1$,
  Dian Chao$^1$,
  Zhihao Guan$^1$,
  Yang Yang\thanks{Corresponding author: Yang Yang(yyang@njust.edu.cn)} $^1$
}

\affil{
  $^1$Nanjing University of Science and Technology\\
}

\maketitle
\ificcvfinal\thispagestyle{empty}\fi

\begin{abstract}

  In this paper, we introduce a grounded video question-answering solution. Our research reveals that the fixed official baseline method for video question answering involves two main steps: visual grounding and object tracking. However, a significant challenge emerges during the initial step, where selected frames may lack clearly identifiable target objects. Furthermore, single images cannot address questions like "Track the container from which the person pours the first time." To tackle this issue, we propose an alternative two-stage approach: (1) First, we leverage the VALOR~\cite{valor,yang2018complex,yang2019comprehensive} model to answer questions based on video information.  (2) concatenate the answered questions with their respective answers. Finally, we employ TubeDETR~\cite{tubedetr,yang2019semi,yy2022} to generate bounding boxes for the targets.
\end{abstract}

\section{Introduction}

Grounded Video Question Answer is a critical task in the field of multimodal understanding. It aims to pinpoint the spatial location of the target in a video based on a posed question, a topic that has garnered increasing attention in recent years. Previous works have primarily focused on question-answering localization for individual images, which has achieved significant progress. However, the domain of videos has not been thoroughly explored. As shown in Figure 1: the competition task calls for generating a series of target boxes for the answers to a given untrimmed video and textual questions. This is a rather challenging task, as it requires the model not only to locate the answer based on the question but also to track the answer.

To achieve better results in grounded video question answering, we explored the traditional two-stage visual question-answering localization methods. As expected, the performance was not satisfactory. Therefore, we propose an alternative two-stage approach: (1) First, we use the VALOR model to answer questions based on video information; (2) concatenate the answered questions with their respective answers, and finally employ TubeDETR to output bounding boxes for the targets.


Our main contributions can be summarized as follows:

\begin{itemize}[itemsep=0pt,parsep=0pt,topsep=0pt,partopsep=0pt,leftmargin=*]
    \item \textbf{Application of TubeDETR: }We conducted an in-depth study and implemented TubeDETR~\cite{tubedetr,yy2023,yang2019adaptive,yang2018complex} single-stage approach. 
    \item \textbf{Prompt with Video QA: }We generate the model's answer through a Video QA model and input it as a prompt to TubeDETR. 
\end{itemize}

\begin{figure}[t]
\begin{center}
   \includegraphics[width=1.0\linewidth]{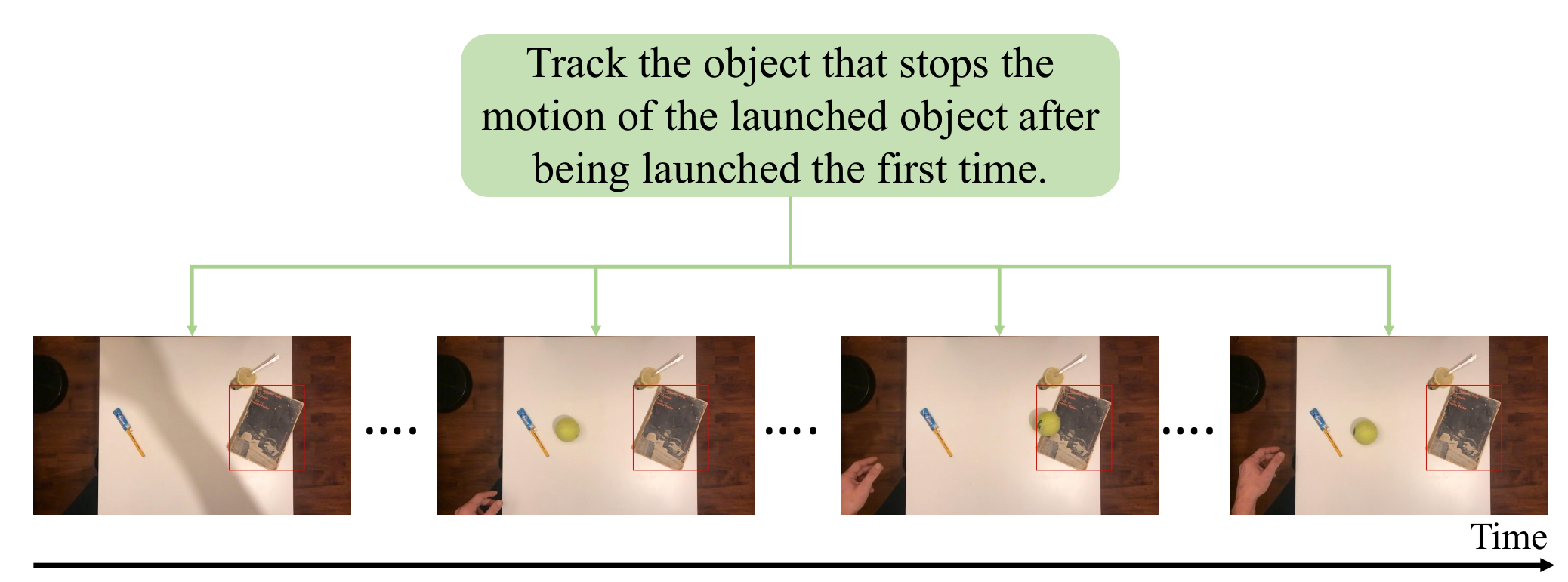}
\end{center}
   \caption{Grounded videoQA needs to answer the question and locate the answer.}
\label{fig:long}
\label{fig:onecol}
\end{figure}

\begin{figure*}
\begin{center}
   \includegraphics[width=1.0\linewidth]{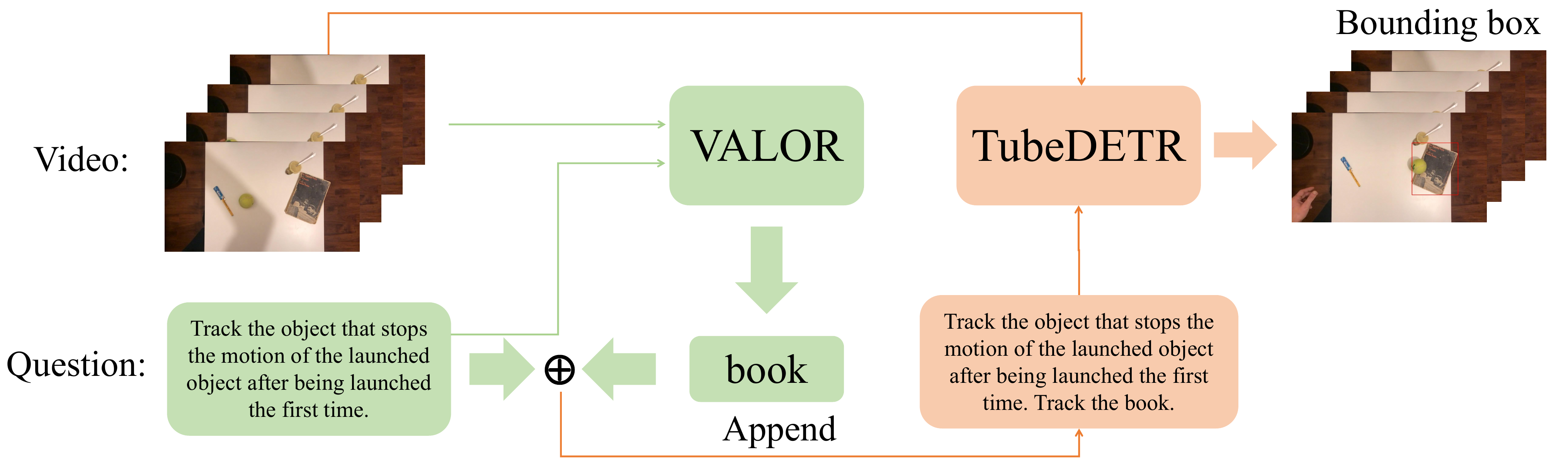}
\end{center}
   \caption{The Overall Architecture of Our Method Based on VALOR and TubeDETR.}
\label{fig:short}
\end{figure*}

\begin{figure}[t]
\begin{center}
   \includegraphics[width=1.0\linewidth]{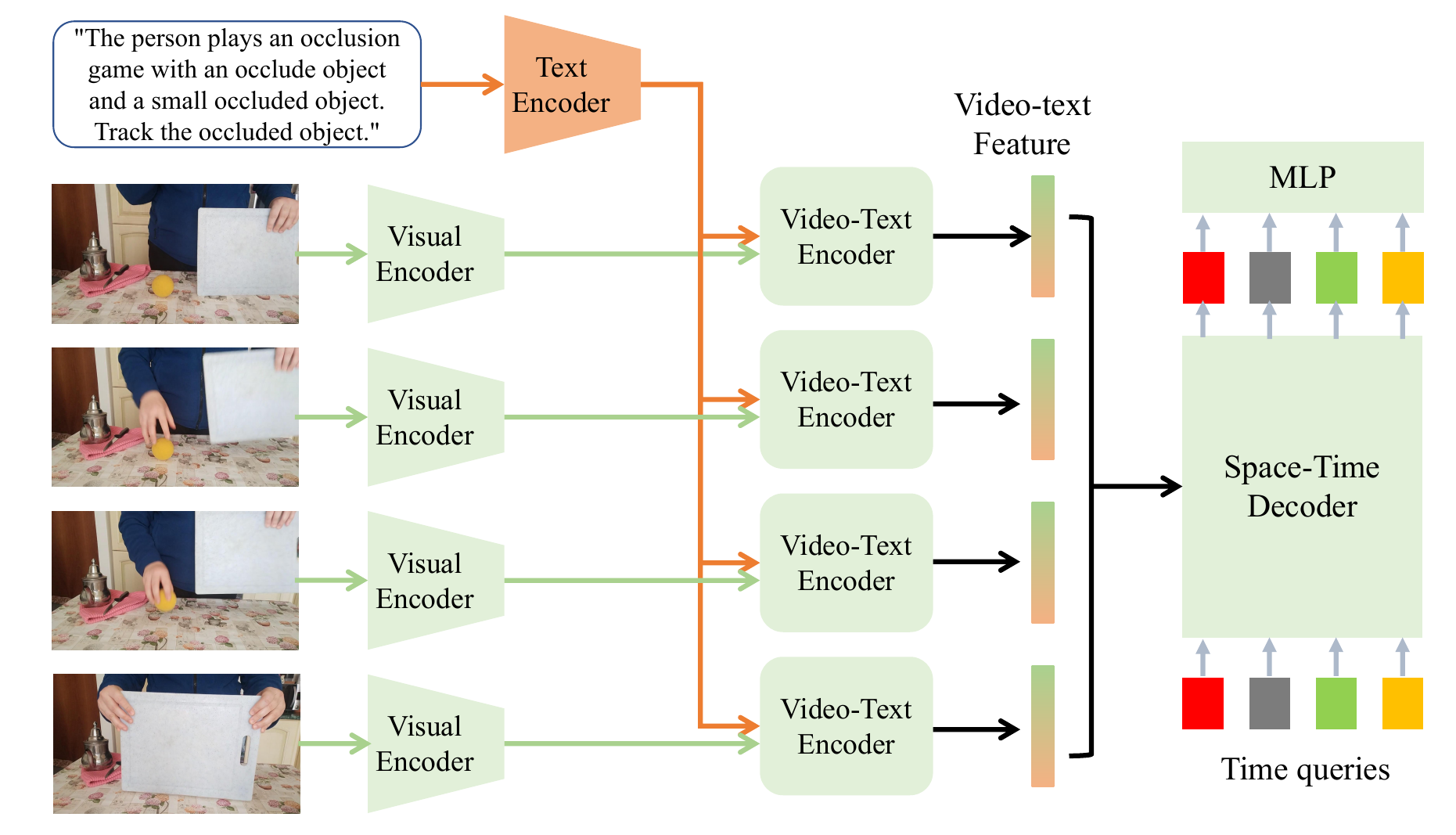}
\end{center}
   \caption{TubeDETR model overview.}
\label{fig:long}
\label{fig:onecol}
\end{figure}


\section{Related Work}

\textbf{Spatio-temporal video grounding.} Spatio-temporal video grounding is a crucial task in the multimodal understanding domain. Its aim is to determine the spatial location of objects in a video based on posed questions, which has garnered increasing attention in recent years. Traditional approaches typically begin by extracting the initial bounding boxes of objects. However, there have been recent developments in one-stage methods, such as TubeDETR~\cite{tubedetr,yang2021corporate} and STVGFormer, which offer new approaches to addressing this challenge.

\textbf{Video Question Answer.} Video Question Answering is one typical type of visual-language task that has been studied for many years. Methodologically, early work proposed various models based on LSTM or graph neural networks to capture cross-modal interactions. Additionally, with the tremendous success of pre-trained visual-language Transformers, VIOLET~\cite{VIOLET} directly fine-tune pre-trained models on downstream Video QA tasks.




\section{Method}
\subsection{Overall Architecture}

We observed that the questions in the dataset primarily revolve around tracking the $k$-th object rather than providing descriptive text about the target object, as is conventionally done in traditional settings. Consequently, it is imperative for the model to demonstrate robust reasoning abilities. Conventional Spatio-temporal video grounding models may not inherently possess this capacity. Therefore, as depicted in Figure 2, we opted to implement a two-stage training approach: (1) Video question answering, and (2) Grounded Video question with answer.



\subsection{Video Question Answer}

Firstly, we utilized the videos, questions, and answers as training data from the training set and employed the VQA task to train the VALOR model. VALOR, a multimodal pre-trained base model, has demonstrated significant performance across various multimodal tasks, which possesses a robust capability to model long videos. In our approach, the input consists of both video and textual questions, with the corresponding answers as output. By integrating visual and textual information, the VALOR model effectively comprehends and responds to questions involving multimodal inputs.

\subsection{Grounded Video question with answer}


As depicted in Figure 2, during the "Grounded Video Question with Answer" phase, we combine the questions from the training set with their respective answer in the format of "{question} Track the {answer}". TubeDETR processes video frames as visual input and treats questions and answers as textual input, ultimately generating bounding boxes for the corresponding objects.

As shown in Figure 3: first, the video frames and question-answer pairs undergo processing through dedicated visual and text encoders, extracting respective visual and textual features. Subsequently, a visual-textual encoder facilitates multimodal interactions. Finally, a spatial-visual decoder utilizes the visual and textual features to generate bounding boxes.

During the inference phase, since the test set does not provide answers corresponding to the questions, we append the answers generated by VALOR and subsequently generate bounding boxes using TubeDETR. Additionally, the competition mandates the provision of boundary predictions for every frame. Given that our videos can contain up to 1000 frames, TubeDETR faces limitations when modeling such long sequences, with a maximum capacity of 200 frames. To overcome this limitation, we sample frames at a rate of 5 frames per second (fps). This sampling method allows us to effectively process video data while retaining crucial information. We then apply TubeDETR to generate predictions based on this sampled data. Finally, we duplicated each predicted bounding box six times to form the final result.

\subsection{EMA}

In our research, we introduced the EMA technique to stabilize the model training process, particularly in the later stages of training. Specifically, we maintain a copy of the EMA model, where its parameters represent the exponentially weighted average of the current model parameters. After each parameter update, we use the following formula to update the parameters of the EMA model:

\begin{align}
    \nu _t = \beta \cdot \nu _{t-1} + (1-\beta)\cdot\theta_t
\end{align}
where $\nu _t$ represents the average of the previous t models, $\beta$ denotes the weighted factor (set to 0.999), and $\theta_t$ signifies the weight of the current model at time t

By employing the EMA method, we are able to maintain a relative stability in the model parameters during the training phase, thereby enhancing the generalization ability of the model and the stability of the training.

\subsection{Weight initialization}

In constructing our architecture, we initialized the model weights using the pre-trained MDETR~\cite{mdetr} models on Flickr30k, MS COCO, and Visual Genome. Additionally, we utilized the pre-trained RoBERTa~\cite{roberta} model weights as the text encoder, and employed the pre-trained ResNet-101 model weights as the image encoder.

\section{Experiment}

\textbf{Dataset.} The dataset is provided by the official competition organizers. Both the training and test sets contain 1859 questions each, while the validation set contains 3051 questions. Specifically, each sample comprises one video and one textual question.

\textbf{Implementation Detail.} In our study, we trained the TubeDETR pre-trained model using an A6000 GPU. The training was conducted for 20 epochs with a learning rate set to 5e-5.

\textbf{Result.} In the results section of our grounded videoQA paper, our findings are presented in Table 1. Here, "Baseline" refers to the official baseline results. "MDETR+Mixform" represents the outcomes of the two-stage approach combining MDETR with Mixform. "VALOR+TubeDETR"  indicates that we first use VALOR to answer questions based on the video, and then employ the TubeDETR model to generate bounding box results.

\begin{table}
\centering
\begin{tabular}{cc}
\toprule
Method & HOTA \\
\hline
MDETR+Mixform  &  0.02\\
Baseline &  0.05\\
VALOR+TubeDETR & 0.06 \\
\toprule
\end{tabular}
\caption{We report the HOTA score of our methods on the test set.}
\end{table}

\section{Conclusion}

The report provides a summary of our solution for the ICCV 2023 Perception Test Challenge 2023 - Task 6 - Grounded VideoQA. In this paper, we attempted TubeDETR  with VALOR to address the Grounded VideoQA task, ultimately surpassing the static baseline in our results.

{\small
\bibliographystyle{ieee_fullname}
\bibliography{main}
}
\end{document}